\def\year{2018}\relax
\pdfoutput=1
\documentclass[letterpaper]{article} 
\usepackage{preprintaaai18}  
\usepackage{times}  
\usepackage{helvet}  
\usepackage{courier}  
\usepackage{url}  
\usepackage[pdftex]{graphicx}  
\usepackage{enumitem}
\usepackage{amsmath}
\usepackage{amssymb}
\begin{document}
%
\title{Video-based Sign Language Recognition without Temporal Segmentation}
%
\author{Jie Huang$^{1}$, Wengang Zhou$^{2}$, Qilin Zhang$^{3}$,  Houqiang Li$^{4}$, Weiping Li$^{5}$\\ 
	$^{1,2,4,5}$Department of Electronic Engineering and Information Science, University of Science and Technology of China\\
	$^3$HERE Technologies, Chicago, Illinois, USA\\
	$^1$hagjie@mail.ustc.edu.cn, \{$^2$zhwg,$^4$lihq,$^5$wpli\}@ustc.edu.cn, $^3$samqzhang@gmail.com}

\maketitle
\begin{abstract}
Millions of hearing impaired people around the world routinely use some variants of sign languages to communicate, thus the automatic translation of a sign language is meaningful and important. Currently, there are two sub-problems in Sign Language Recognition (SLR), i.e., isolated SLR that recognizes word by word and continuous SLR that translates entire sentences. Existing continuous SLR methods typically utilize isolated SLRs as building blocks, with an extra layer of preprocessing (temporal segmentation) and another layer of post-processing (sentence synthesis). Unfortunately, temporal segmentation itself is non-trivial and inevitably propagates errors into subsequent steps. Worse still, isolated SLR methods typically require strenuous labeling of each word separately in a sentence, severely limiting the amount of attainable training data. To address these challenges, we propose a novel continuous sign recognition framework, the Hierarchical Attention Network with Latent Space (LS-HAN), which eliminates the preprocessing of temporal segmentation. The proposed LS-HAN consists of three components: a two-stream Convolutional Neural Network (CNN) for video feature representation generation, a Latent Space (LS) for semantic gap bridging, and a Hierarchical Attention Network (HAN) for latent space based recognition. Experiments are carried out on two large scale datasets. Experimental results demonstrate the effectiveness of the proposed framework.
\end{abstract}

\section{Introduction}
A key challenge in Sign Language Recognition (SLR) is the design of visual descriptors that reliably captures body motions, gestures, and facial expressions. There are primarily two categories: the hand-crafted features \cite{sun2013discriminative,koller2015continuous} and Convolutional Neural Network (CNN) based features \cite{tang2015real,huang2015sign,pu2016sign}. Inspired by the recent success in CNN \cite{s17102421,s17061341}, we design a two-stream 3D-CNN for video feature extraction. 

Temporal segmentation is also a difficulty in continuous Sign Language Recognition (SLR). The common scheme to continuous SLR is to decompose it to isolated word recognition problem, which involves temporal segmentation. Temporal segmentation is non-trivial since the transitional movements are diverse and hard to detect, and as a preprocessing step, inaccurate segmentation could incur errors in the subsequent steps. In addition, it is highly time consuming to label each isolated fragments.

Motivated by video caption with Long-Short Term Memory (LSTM), we circumvent the temporal segmentation with Hierarchical Attention Network (HAN), an extension to LSTM by considering structure information and attention mechanism. The scheme is to feed HAN with the entire video and output complete sentence word-by-word. However, HAN locally optimizes the probability of generating the next word given the input video and previous word, ignoring the relationship between video and sentences \cite{pan2015jointly}. As a result, it could suffer from robustness issues. To remedy this, we incorporate a Latent Space model to explicitly exploit the relationship between visual video and text sentence.


In summary, the major contributions of the paper are:
\begin{itemize}[leftmargin=*,itemsep=0cm,topsep=0cm,parsep=0cm]
	\item A new two-stream 3D CNN for the generation of global-local video feature representations;
	\item A new LS-HAN framework for continuous SLR without requiring temporal segmentation;
	\item Joint optimization of relevance and recognition loss in the proposed LS-HAN framework; 
	\item Compilation of the largest (as of September 2017) open-source Modern Chinese Sign Language (CSL) dataset for continuous SLR with sentence-level annotations.  
\end{itemize}

\section{Related Work}\label{sec1}
In this section, a brief review of continuous SLR, video subtitle generation, and latent space model is given.

\subsection{Continuous SLR}
Most existing SLR researches \cite{huang2015sign,guo2016sign,guo2016online,liu2016sign} fall into the category of isolated SLR, i.e., recognition of words or expressions, similar to action recognition \cite{cai2016effective}. A more challenging problem is the continuous SLR, which involves the reconstruction of sentence structures. Most existing continuous SLR methods divide the sentence-to-sentence recognition problem into three stages, temporal segmentation of videos, isolated word/expression recognition (i.e., isolated SLR), and sentence synthesis with a language model. For example, DTW-HMM \cite{zhang2014threshold} proposed a threshold matrix based coarse temporal segmentation step followed by a Dynamic Time Warping (DTW) algorithm and a bi-grammar model. In \cite{Koller_2017_CVPR}, a new HMM based  language model is incorporated. Recently, transitional movements attract a lot of attention \cite{dawod2016gesture,li2016sign,yang2016continuous,zhang2016chinese} because they can serve as the basis for temporal segmentation. 

Despite its popularity, temporal segmentation is intrinsically difficult: even the transitional movements between hand gestures can be subtle and ambiguous. Inaccurate segmentation can incur significant performance penalty on subsequent steps \cite{zhang2014threshold,koller2015continuous,fang2007large}. Worse still, the isolated SLR step typically requires per-video-frame labels, which are highly time consuming.

\subsection{Video Description Generation} %
Video description generation \cite{pan2015jointly,venugopalan2015sequence,yao2015describing} is a relevant research area, which generates a brief sentence describing the scenes/objects/motions of a given video sequences. One popular method is the sequence-to-sequence video-to-text \cite{venugopalan2015sequence}, a two layer LSTM on top of a CNN.  Attention mechanism can be incorporated into a LSTM \cite{yao2015describing}, which automatically selects the most likely video frames. There are also some extensions to the LSTM-style algorithms, such as bidirectional LSTM \cite{bin2016bidirectional}, hierarchical LSTM \cite{li2015hierarchical} and hierarchical attention GRU \cite{yang2016hierarchical}.

Despite many similarities in targets and involved techniques, video description generation and continuous SLR are two distinctive tasks. Video description generation provides a brief summary of the appearance of a video sequence, such as ``A female person raises a hand and stretch fingers" while the continuous SLR provides a semantic translation of sign language sentences, such as ``A signer says `I love you' in American Sign Language. '' 

\subsection{Latent Space based Learning} 
Latent space model is a popular tool to bridge the semantic gap between two modalities \cite{zhang2015can,zhang2015auxiliary,zhang2015multi}, such as between textual short descriptions and imagery \cite{pan2014click}. For example, the combination of LSTM and latent space model is proposed in \cite{pan2015jointly}  for video caption generation, in which both an embedding model and an LSTM are jointly learned. However, this embedding model is purely based on the Euclidean distance between video frames and textual descriptions, ignoring all temporal structures. It is suitable for the generation of video appearance summaries, but falls short of semantic translation.
\section{Signing Video Feature Representation}
\label{sec:rep}
Signing video is characterized primarily by upper body motions, especially hand gestures. The major challenge of hand gesture detection and tracking is the tremendous appearance variations of hand shapes and orientations, in addition to occlusions. Previous researches \cite{tang2015real,kurakin2012real,sun2013discriminative} on posture/gesture models rely heavily on collected depth maps from Kinect-style RGB-D sensors, which provides a convenient way in 3D modeling. However, a vast majority of existing annotated signing videos are recorded with conventional RGB camcorders.  An effective RGB video based feature representation is essential to take advantage of these valuable historical labeled data.

Inspired by the recent success of deep learning based object detection, we propose a two-stream 3-D CNN for the generation of video feature representation, with accurate gesture detection and tracking. This particular two-stream 3-D CNN takes both the entire video frames and cropped/tracked gesture image patches as two separate inputs for each stream, with a late fusion mechanism achieved through shared final fully connected layers. Therefore, the proposed CNN encodes both global and local information. 

\subsection{Gesture Detection and Tracking}
Faster R-CNN \cite{girshick2015fast} is a popular object detection method, which is adopted in our proposed method for gesture detection and tracking. We pre-train a faster R-CNN on the VOC2007\footnote{http://host.robots.ox.ac.uk/pascal/VOC/voc2007/index.html} person-layout dataset\footnote{It contains detailed body part labels (\textit{e.g}., head, hands and feet).} with two output units representing hands versus background. Subsequently, we randomly select $400$ frames from our proposed CSL dataset and manually annotate the hand locations for fine-tuning. After that, each video is processed frame-by-frame for gesture detection. 
%

The faster R-CNN based detection can fail when the hand-shape varies hugely or is occluded by clothes. To localize gestures in these frames, compressive tracking \cite{zhang2012real} is utilized. The compressive tracking model represents target regions in multi-scale compressive feature vectors and scores the proposal regions with a Bayes classifier. The Bayes classifier parameters are updated based on detected targets in each frame, the compressive tracking model is robust to huge appearance variations. Specifically, a compressive tracking model is initialized whenever the faster R-CNN detection fails, with the successfully detections from the immediate prior video frame. Conventional tracking algorithms often suffer from the drift problem, especially with long video sequences. Our proposed method is largely immune to this problem due to limited length of sign videos. 
\subsection{Two-stream 3D CNN}
\begin{figure}[t]
\begin{center}
\includegraphics[width=0.9\linewidth]{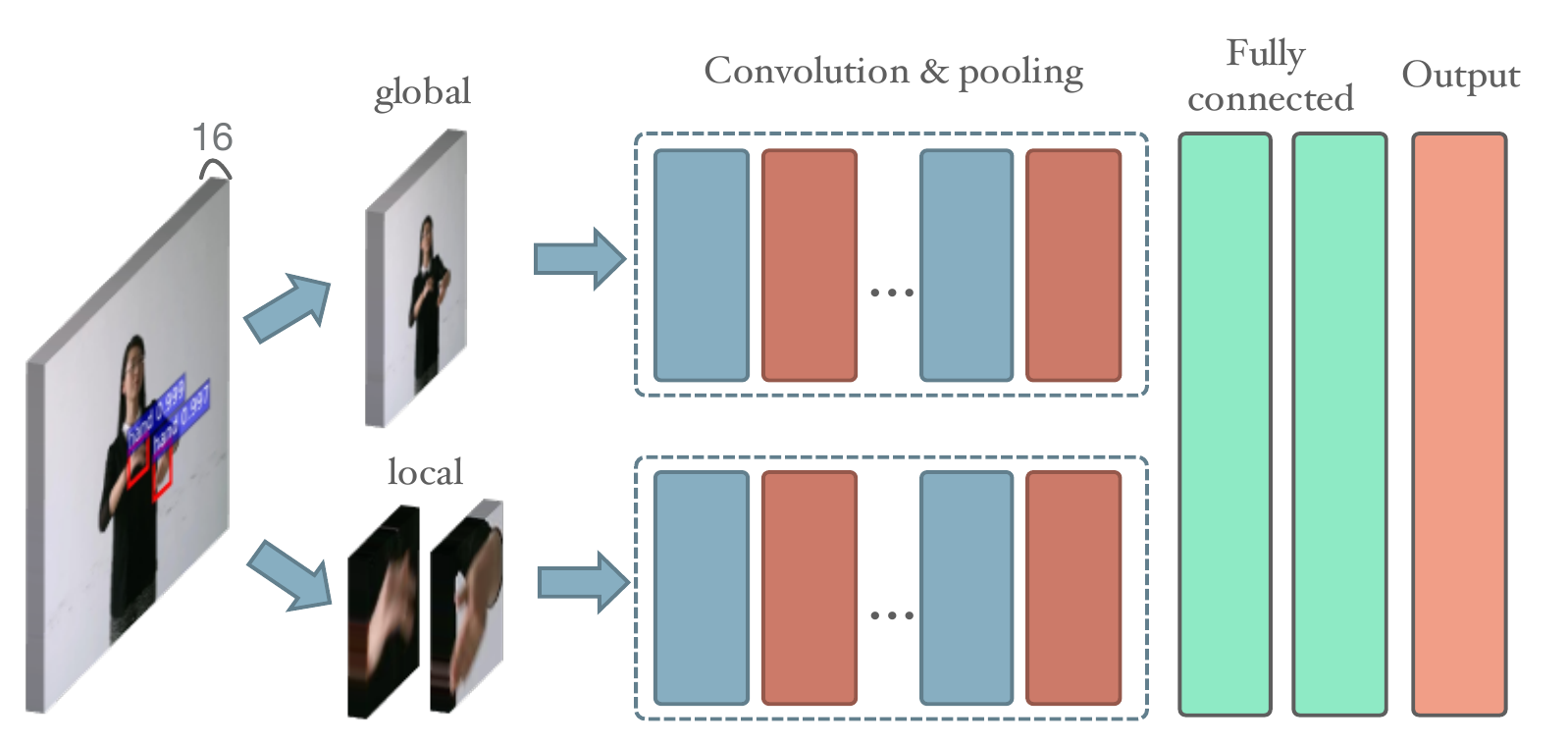}
\end{center}
\caption{Two-stream 3-D CNN. Input is a video clip containing 16 adjacent frames. Two streams share the same structure as C3D~\cite{tran2015learning}. Blue and red blocks represent convolutional and pooling layers, respectively. Global-local feature representations are fused in the rightmost two fully connected layers.}
\label{fig:twostream}
\end{figure}

Due to the nature of signing videos, a robust video feature representation requires the incorporation of both global hand locations/motions and local hand gestures. As shown in Fig.~\ref{fig:twostream}, we design a two-stream 3-D CNN based on the C3D~\cite{tran2015learning}, which extracts spatio-temporal features from a video clip (16 frames as suggested in \cite{tran2015learning}). The input to the network is a video clip containing adjacent 16 frames. The upper stream is designed to extract global hand locations/motions, where the input is resized ($227 \times 227$) complete video frames. The lower stream focuses on the local, detailed hand gestures, where the input is cropped (also $227 \times 227$), tracked image patches containing tight bounding boxes of hands. Left and right hand patches are concatenated as multi-channel inputs. Each stream shares the same network structure as the C3D network, including eight convolutional layers and five pooling layers. Two fully connected layers serve as fusion layers to combine global and local information from the upper and lower stream, respectively. 

The two-stream CNN is first pre-trained with an isolated SLR dataset \cite{pu2016sign}, after which all weights are fixed and the last SoftMax and fully-connected layer are discarded. During the test phase, this CNN is truncated at the first fully connected layer (as a feature extractor) and each test video is divided into 16-frame clips with a temporal sliding window and subsequently fed into the two-steam 3D CNN. The $4096$ dimensional output of the truncated 3D CNN is the desired global-local feature representation for this particular 16-frame video clip. Consequently, each video is denoted as a sequence of such $4096$ dimensional feature vectors.
\section{Proposed LS-HAN Model}
\label{sec:mod}
In this section, we present the Hierarchical Attention Network (HAN) for continuous SLR in a latent space. HAN is an extension to LSTM, which incorporates the attention mechanism based on the structure of input. 
In the proposed joint learning model, the optimization function takes into account both the video-sentence relevance error $E_{r}$ in a latent space, and a recognition error $E_{c}$ by HAN, 
\begin{equation}\label{eq:loss0}
\begin{aligned}
\underset{\theta_{r},\theta_{c}}{min}\frac{1}{N}  \sum_{i=1}^{N} & \lambda_1E_{r}(V^{(i)},S^{(i)};\theta_{r}) \\
& +(1-\lambda_1)E_{c}(V^{(i)},S^{(i)};\theta_{r},\theta_{c}) + \lambda_2R~,
\end{aligned}
\end{equation}
where $N$ denotes the number of instances in the training set, and $i_{th}$ instance being a video with annotated sentence $(V^{(i)},S^{(i)})$. $\theta_{r}$ and $\theta_{c}$ denote parameters in the latent space and HAN, respectively. $R$ is a regularization term. Eq.~\eqref{eq:loss0} represents the minimization of the mean loss over training data with some regularizations. The balance between the loss term and the regularization term is achieved by weights $\lambda_1$ and $\lambda_2$.

In the following, the term $E_{r}$ is first formulated in the video-sentence latent space, followed by details on sentence recognition with HAN and the formula of term $E_{c}$. Finally, the overall training and testing processes of LS-HAN are presented.
\begin{figure}[t]
\begin{center}
\includegraphics[width=0.95\linewidth]{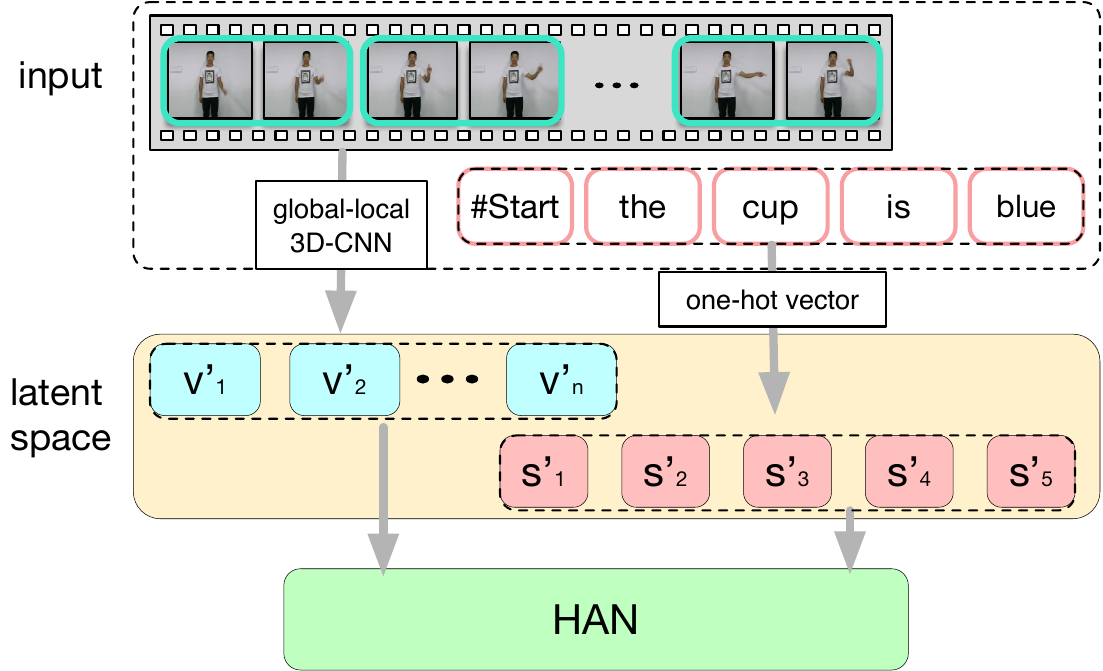}
\end{center}
\caption{The proposed LS-HAN framework. Input are video paired with annotation sentence. Video is represented with global-local features and each word is encoded with one-hot vector. They are mapped into the same latent space to model video-sentence relevance. Based on the mapping results, we utilize HAN for automatic sentence generation.}
\label{fig:model}
\end{figure}

\subsection{Video-sentence Latent Space}
\label{sec:ls}

As shown in Fig.~\ref{fig:model}, the input to our framework is videos with paired annotated sentences. Video is represented with the proposed global-local features, while each word of the annotated sentence is encoded with a ``one-hot'' vector\footnote{A one-hot vector is a binary index used to distinguish words with a given vocabulary. All bits are `0', except the $i$th bit being `1' for the $i$th word in this vocabulary.}.

Let the videos be $V=(v_1, v_2, ..., v_n)$, and the sentence be $S=(s_1, s_2, ..., s_m)$, where $v_i\in \mathbb{R}^{D_c}$ is the $D_c$-dimensional global-local feature of the $i_{th}$ video clip. $n$ denotes the total number of clips in this video, $s_j\in \mathbb{R}^{D_w}$ is the ``one-hot'' vector of $j_{th}$ word of sentence, $D_w$ is vocabulary size and $m$ represents the total number of words in the current sentence.

The goal of the latent space is to construct a space to bridge semantic gaps, we map $V\in \mathbb{R}^{D_c\times n}$ and $S\in \mathbb{R}^{D_w\times m}$ to the same latent space as $f_v(V)=(v'_1, v'_2, ..., v'_n)$ and $f_s(S)=(s'_1, s'_2, ..., s'_m)$, where $f_v$ and $f_s$ are mapping function for video features and sentence features, respectively,
\begin{equation}\label{eq:map}
f_v(x) = T_vx  \quad \text{and} \quad f_s(x) = T_sx ~,
\end{equation}
$T_v\in \mathbb{R}^{D_{s}\times D_c}$ and $T_s\in \mathbb{R}^{D_{s}\times D_w}$ are transformation matrices that project video content and semantic sentence into common space. $D_s$ is the dimension of the latent space.

To measure the relevance between $f_v(V)$ and $f_s(S)$, the Dynamic Time Warping (DTW) algorithm is used to find the minimal accumulating distance of two sequences and the temporal warping path, 
\begin{equation}\label{eq:dtw1}
D[i, j]=min(D[i-1, j], D[i-1,j-1]) + d(i, j)~,
\end{equation}
\begin{equation}\label{eq:eu}
d(i, j) =||T_vv_i - T_ss_j||_2~,
\end{equation}
where $D[i, j]$ denotes the distance between $(v'_1, ..., v'_i)$ and $(s'_1, ..., s'_j)$, and $d(i,j)$ denotes the distance between $v'_i$ and $s'_j$. Thus we define the loss of video-sentence with single instance as,
\begin{equation}\label{eq:els}
E_{r}(V,S;\theta_{r}) = D(n, m)~.
\end{equation}

This DTW algorithm assumes that words appearing earlier in a sentence should also appear early in video clips, i.e., a monotonically increasing alignment. Per our evaluation dataset, most signers vastly prefer simple sentences to compound ones, thus simple sentences with a single clause makes up the majority of the dataset. Therefore, approximate monotonically increasing alignment can be assumed.

\begin{figure}[t]
\begin{center}
\includegraphics[width=0.9\linewidth]{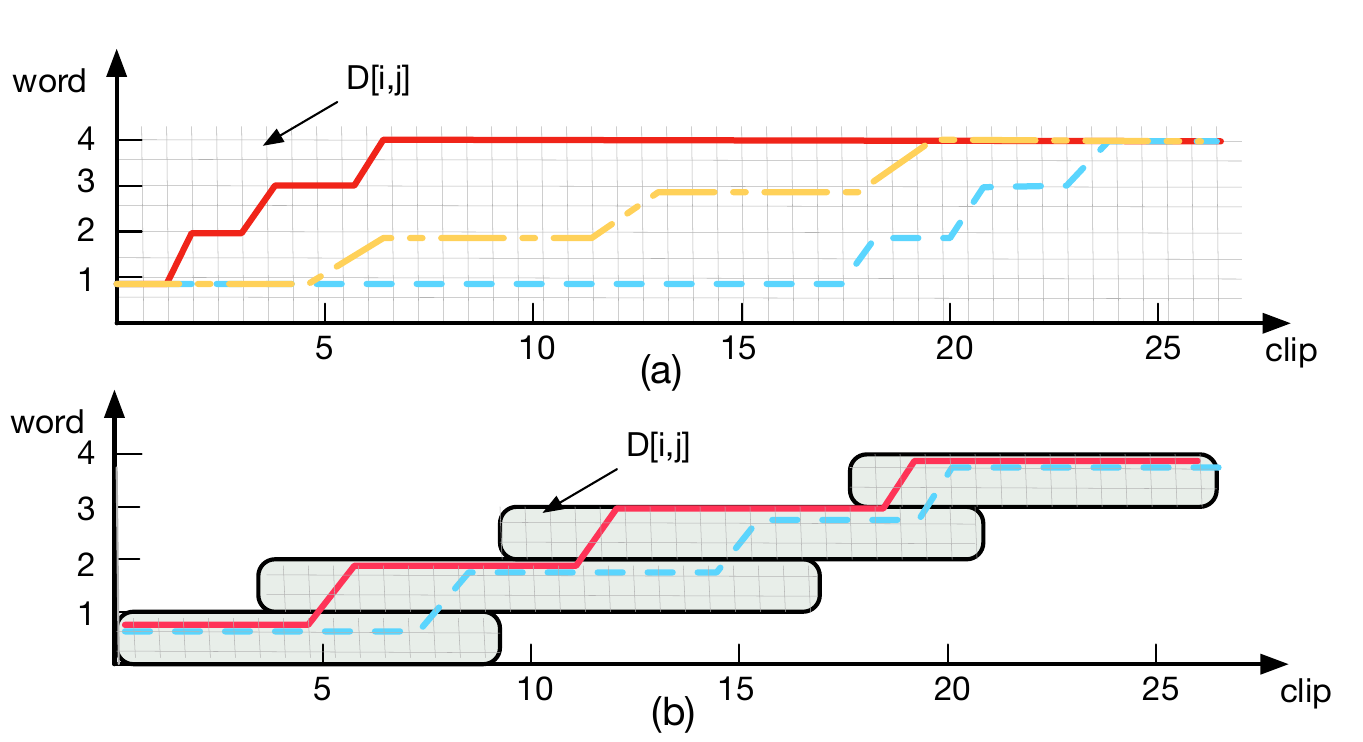}
\end{center}
\caption{Associated warping paths generated by DTW. X-axis represents the frame index, y-axis indicates the word sequence index. Grids represent the matrix elements $D[i,j]$. (a) shows three possible alignment paths of raw DTW. (b) shows the alignment paths of Window-DTW which are bounded by windows.}
\label{fig:dtw}
\end{figure}

The associated warping path is recovered using back-tracking. This warping path is normally interpreted as the alignment between frames and words. For better alignment, the Windowing-DTW \cite{biba2011learning} is used, as shown in Fig.~\ref{fig:dtw}, and $D[i,j]$ is only computed within the windows. In the following experiments, the window length is $\frac{n}{2}$, except the first and last one at boundary $0$ and $n$. All adjacent windows have $\frac{n}{4}$ overlap. 

\subsection{Recognition with HAN}
\label{sec:lstm}
Inspired by recent sequence-to-sequence models \cite{venugopalan2015sequence}, the recognition problem is formulated as the estimation of log-conditional probability of sentences given videos. By minimizing the loss,  contextual relationship among the words in sentences can be kept. 
\begin{equation}\label{eq:etr1}
\begin{aligned}
E_{c}(V, S;\theta_{r},\theta_{c}) = &\log p(s'_1,s'_2,...,s'_m|v'_1,v'_2,...,v'_n;\theta_{c}) \\
=& \sum_{t=1}^{m}\log p(s'_t | v'_1, ..., v'_n,s'_1, ..., s'_{t-1};\theta_{c})~.
\end{aligned}
\end{equation}
First, input frame sequences are encoded as a latent vector representation on a per-frame basis, followed by decoding from each representation to a sentence, one word at a time. We extend the encoder in HAN to reflect the hierarchical structures in its inputs (clips form word and words form sentence) and incorporate the attention mechanism. The modified model is an unnamed variant of HAN. 
\begin{figure}[t]
\begin{center}
\includegraphics[width=0.95\linewidth]{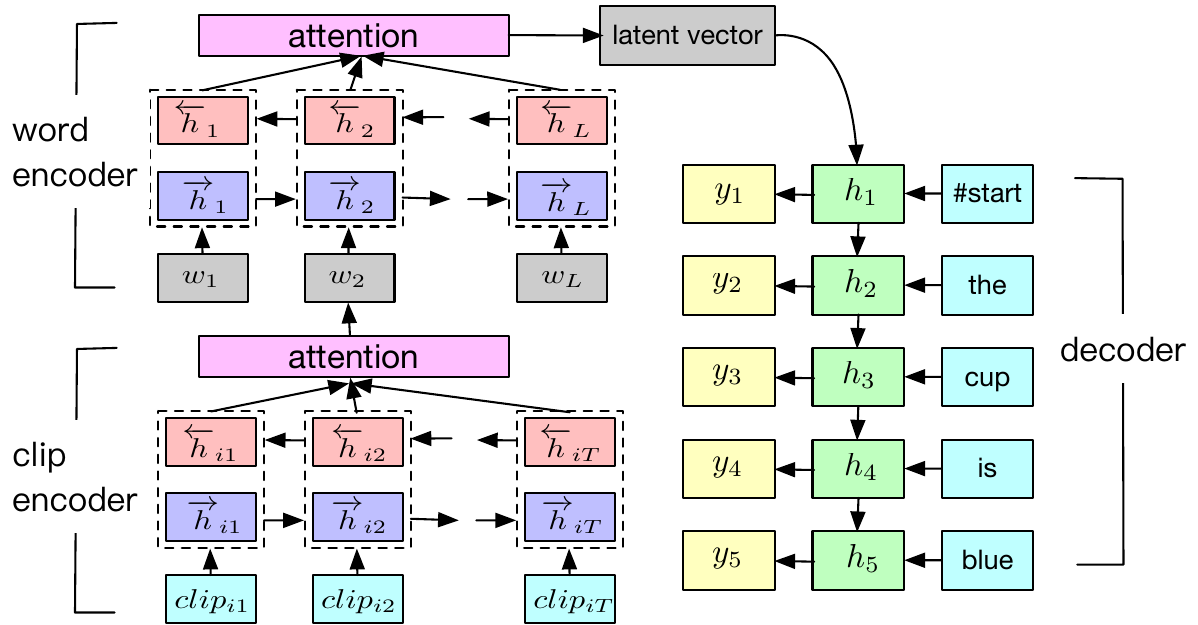}
\end{center}
\caption{HAN encodes videos hierarchically and weights the input sequence via a attention layer. It decodes the hidden vector representation to a sentence word-by-word. }
\label{fig:han}
\end{figure}

\begin{figure}[t]
\begin{center}
\includegraphics[width=0.95\linewidth]{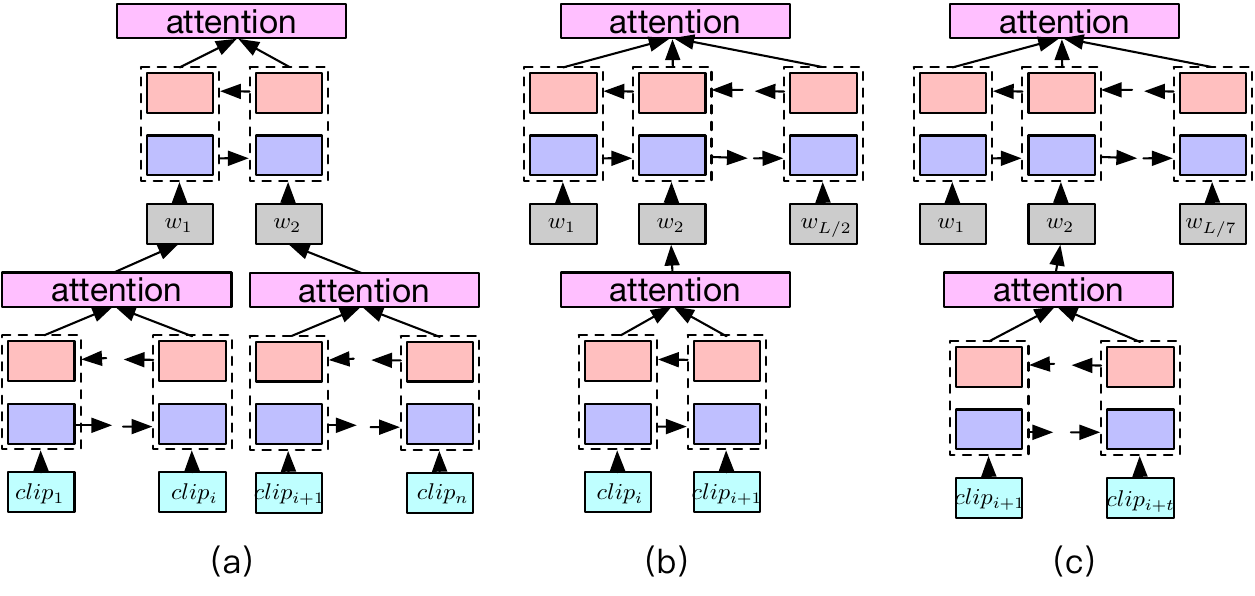}
\end{center}
\caption{Alignment reconstruction in the testing phase. (a) Split clips to two subsequences and encode as HAN; (b) Split to subsequences every two adjacent clips; (c) Split evenly to $7$ subsequences, where $7$ is average sentence length in the training set.}
\label{fig:han-recog}
\end{figure}

As shown in Fig.~\ref{fig:han}, the inputs (in blue) are clip sequences and word sequences represented in latent space. This model contains two encoders and a decoder. Each encoder is a bidirectional LSTM with a attention layer; while the decoder is a single LSTM. The clip encoder encodes the video clips aligning to a word. As shown in Fig.~\ref{fig:han-recog}, we empirically try three alignment strategies:
\begin{enumerate}[noitemsep,leftmargin=*]
	\item Split clips to two subsequences;
	\item Split to subsequences every two clips;
	\item Split evenly to $7$ subsequences, since each sentence contains 7 words on average in the training set.
\end{enumerate}
The outputs of bidirectional LSTM pass through the attention layer to form a word-level vector, where the attention layer acts as information selecting weights to the inputs. Subsequently, the sequences of word-level vectors are projected into a latent vector representation, where decoding is carried out with annotation sentence. During the decoding, $\#Start$ is used as the start symbol to mark the beginning of a sentence and $\#End$ as the end symbol that indicates the end of a sentence. At each time stamp, words encoded with ``one-hot vector'' are mapped into the latent space and fed to the LSTM (denoted in green) together with the hidden state from the previous timestamp. The LSTM cell output $h_t$ is used to emitted word $y_t$. A softmax function is applied to obtain the probability distribution over the words $\hat{y}$ in the vocabulary $voc$. 
\begin{equation}\label{eq:prob}
p(y_t|h_t)=\frac{\exp(W_{y_t}h_t)}{\sum_{\hat{y}\in voc}\exp(W_{\hat{y}}h_t)}, 
\end{equation}
where $W_{y_t}, W_{\hat{y}}$ are parameter vectors in the softmax layer.  All next words are obtained based on the probability in Eq.~\eqref{eq:prob} until the sentence end symbol is emitted. 

Inspired by \cite{yang2016hierarchical}, the coherence loss in Eq.~\eqref{eq:etr1} can be further simplified as, 
\begin{equation}\label{eq:etr2}
\begin{aligned}
E_{c}(V, S;\theta_{r},\theta_{c}) = & \sum_{t=1}^{m}\log p(s'_t | h_{t-1},  s'_{t-1}) \\
=& \sum_{t=1}^{m} \log \frac{\exp(W_{s_t}h_t)}{\sum_{\hat{s}'\in voc}\exp(W_{\hat{s}'}h_t)}~.
\end{aligned}
\end{equation}
%
\subsection{Learning and Recognition of LS-HAN Model}\label{subsec:lr}
With previous formulations of the latent space and HAN, Eq.~\eqref{eq:loss0} is equivalent to, 
\begin{equation}\label{eq:loss}
\begin{aligned}
\underset{T_v,T_s,\theta }{min}\frac{1}{N}\sum_{i=1}^{N}&\lambda_1E_{r}(V^{(i)},S^{(i)};T_v,T_s) \\
& +(1-\lambda_1)E_{c}(V^{(i)},S^{(i)};T_v,T_s,\theta)  \\
& +\lambda_2 (\left \| T_v \right \|^2 + \left \| T_s \right \|^2+ \left \| \theta \right \|^2)~,
\end{aligned}
\end{equation}
where $T_v$ and $T_s$ are latent space parameters, $\theta$ is a HAN parameter. The optimization of LS-HAN requires the partial derivatives of Eq.~\eqref{eq:loss} with respect to parameters $T_v$, $T_s$ and $\theta$ and the solutions of such equations. Thanks to the linearity, the two loss terms and the regularization item could be optimized separately. For simplicity, we only elaborate the computation of partial derivatives of $E_{r}$ and $E_{c}$ in the simplest case (with only one sample), ignoring unrelated coefficients and regularization terms.

Consider the partial derivative of $E_r$ with respect to matrices $T_v$ and $T_s$. According to Eq.~\eqref{eq:els}, 
\begin{equation}
\frac{\partial E_{r}(V,S;T_v,T_s)}{\partial T_v}=\frac{\partial D(n, m)}{\partial T_v}~, \label{partialENewlabel}
\end{equation}
Substitution of Eq.~\eqref{eq:dtw1} in Eq.~\eqref{partialENewlabel} leads to 
\begin{equation}\label{partialDnew}
\begin{aligned}
\frac{\partial D(n,m)}{\partial T_v} = &\min(\frac{\partial D(n-1,m)}{\partial T_v},\frac{\partial D(n-1,m-1))}{\partial T_v})\\
&+ \frac{\partial d(n,m)}{\partial T_v}. 
\end{aligned}
\end{equation}
Eq.~\eqref{partialDnew} is recursive and allows for gradient back-propagation through time. As a result, $\frac{\partial D(n,m)}{\partial T_v} $ can be represented by $\frac{\partial d(i,j)}{\partial T_v}$ for $i<n$, $j<m$, 
\begin{equation}
\frac{\partial D(n,m)}{\partial T_v}=f\left(\left \{  \frac{\partial d(i,j)}{\partial T_v} | i<n, j<m\right \}\right)~,
\end{equation}
where $f(\cdot)$ denotes a function that can be represented. Since the term $\frac{\partial d(i,j)}{\partial T_v}$ can be obtained according to Eq.~\eqref{eq:eu}, $\frac{\partial D(n,m)}{\partial T_v} $ can be computed by the chain rule. Likewise, $\frac{\partial E_r(V,S;T_v,T_s)}{\partial T_s}$ could be similarly computed. 

Consider the partial derivative of HAN error $E_c$ with respect to $T_v$, $T_s$ and the network parameter $\theta$. Regular stochastic gradient descent is utilized, with gradients calculated by the Back-propagation Through Time (BPTT) \cite{werbos1990backpropagation}, which recursively back-propagates gradients from current to previous timestamps. After gradients are propagated back to the inputs, $\frac{\partial E_c(V,S;T_v,T_s,\theta)}{\partial V'}$ and $\frac{\partial E_c(V,S;T_v,T_s,\theta)}{\partial S'}$ are obtained. With Eq.~\eqref{eq:map}, further simplification could be obtained, 
\begin{equation}
\frac{\partial E_c(V,S;T_v,T_s,\theta)}{\partial T_v} = \frac{\partial E_c(V,S;T_v,T_s,\theta)}{\partial V'} \cdot V~,
\end{equation}
\begin{equation}
\frac{\partial E_c(V,S;T_v,T_s,\theta)}{\partial T_s} = \frac{\partial E_c(V,S;T_v,T_s,\theta)}{\partial S'} \cdot S~.
\end{equation}
Therefore, all gradients with respect to all parameters could be obtained. The objective function in Eq.~\eqref{eq:loss} can be minimized accordingly. 


During the testing phase, the proposed LS-HAN is used to translate signing videos sentence-by-sentence. Each video is divided into clips with a sliding window algorithm. The alignment information need to be reconstructed. Empirical experiments verify that the strategy 3 outperforms others, therefore it is employed in the testing phase. After encoding, the start symbol ``$\#Start$'' is fed to HAN indicating the beginning of sentence prediction. During each decoding timestamp, the word with the highest probability after the softmax is chosen as the predicted word, with its representation in the latent space fed to HAN for the next timestamp, until the emission of the end symbol ``$\#End$''. 
\section{Experiments}\label{sec:exp}
In this section, datasets are introduced followed by evaluation comparison. Additionally, a sensitivity analysis on a trade-off parameter is included. 
\subsection{Datasets}
Two open source continuous SLR datasets are used in the following experiments, one for CSL and the other is the German sign language dataset RWTH-PHOENIX-Weather \cite{koller2015continuous}. The CSL dataset in Tab.~\ref{tab:dataset} is collected by us and released on our project web page\footnote{\url{https://ustc-slr.github.io/datasets/2015_csl}}. A Microsoft Kinect camera is used for all recording, providing RGB, depth and body joints modalities in all videos. The additional modalities should provide helpful additional information as proven in hyper-spectral imaging efforts \cite{s17102421,zhang2011fast,zhang2012fast,abeida2013iterative}, which is potentially helpful in future works. In this paper, only the RGB modality is used. The CSL dataset contains 25K labeled video instances, with $100+$ hours of total video footage by 50 signers. Every video instance is annotated with a complete sentence by a professional CSL teacher. 17K instances are selected for training, 2K for validation, and the rest 6K for testing. The RWTH-PHOENIX-Weather dataset contains 7K weather forecasts sentences from 9 signers. All videos are of 25 frames per second (FPS) and at resolution of $210\times260$. Following \cite{koller2015continuous}, 5,672 instances are used for training, 540 for validation, and 629 for testing.
\begin{table}
\centering
\footnotesize
\caption{Statistics on Proposed CSL Video Dataset} \label{tab:dataset}
\begin{tabular}{l l|l l}
\hline
RGB resolution & 1920$\times$1080 & \# of signers & 50 \\ \hline
Depth resolution &512$\times$424 & Vocab. size & 178 \\ \hline
Video duration (sec.) & 10$\sim$14 & Body joints & 21 \\ \hline
Average words$/$instance & 7  & FPS & 25 \\ \hline
Total instances & 25,000 & Total hours & 100+\\
\hline
\end{tabular}
\end{table}

\subsection{Experimental Setting} %
Per \cite{tran2015learning}, videos are divided into 16-frame clips with 50\% overlap, with frames cropped and resized at $227\times227$. The outputs of the 4096-dimensinoal $fc6$ layer from 2-stream 3D CNN are clip representations. The following parameters are set based on our validation set. The dimension of latent space and the size of hidden layer in HAN are both 1024. The trade-off parameter $\lambda_1$ in Eq.~\eqref{eq:loss} of relevance loss and coherence loss is set to 0.6. The regularization parameter is empirically set to 0.0002.

\subsection{Evaluation Metrics}%
\label{subsubsec:metrics}
Predicted sentence can suffer from errors including word substitution, insertion and deletion errors, following~\cite{fang2007large,starner1998real,zhang2014threshold},
\begin{equation}\label{eq:accu}
	Accuracy = 1-\frac{S+I+D}{N}\times 100\%~,
\end{equation}
where $S,I,$ and $D$ denote the minimum number of substitution, insertion and deletion operations needed to transform a hypothesized sentence to the ground truth. $N$ is the number of words in ground truth. Note that, since all errors are accounted against the accuracy rate, it is possible to get negative accuracies.

\subsection{Results and Analyses}
%
%
\begin{table}
	\centering
	\small
	\caption{Continuous SLR Results. Methods in bold text are the original and modified versions of the proposed method.} \label{tab:accu}
	\setlength{\tabcolsep}{10pt} %
	\begin{tabular}{l|c} \hline
		Methods &  Accuracy\\ \hline
		LSTM \cite{venugopalan2014translating} &  0.736 \\ 
		S2VT  \cite{venugopalan2015sequence} &  0.745 \\ 
		LSTM-A  \cite{yao2015describing}           &  0.757 \\
		LSTM-E \cite{pan2015jointly}  	               & 0.768 \\ 
		HAN \cite{yang2016hierarchical}  	       & 0.793 \\ \hline
		CRF \cite{lafferty2001conditional}            & 0.686 \\
		LDCRF \cite{morency2007latent}		& 0.704 \\
		DTW-HMM \cite{zhang2014threshold}	& 0.716 \\ \hline
		\textbf{LS-HAN (a)} &  0.792\\
		\textbf{LS-HAN (b)} &  0.805\\
		\textbf{LS-HAN (c)} &  \textbf{0.827}\\
		\hline
	\end{tabular}
\end{table}
The continuous SLR results over our CSL dataset is summarized in Tab.~\ref{tab:accu}. Since our framework is based on LSTM, we compare with LSTM, S2VT, LSTM-A, LSTM-E and HAN, which are also related to LSTM. Note that LSTM-E that jointly learns LSTM and embedding layer is much similar to our method. One of the key differences is that LSTM-E ignores temporal information for brevity during its embedding process, while we choose to retain temporal structural information while optimizing video-sentence correspondence. Given the result that our method achieve 0.059 higher accuracy than LSTM-E, our solution to model video-sentence relevance is a better choice.

Besides, we also make a comparison with  continuous SLR algorithms: CRF, LDCRF and DTW-HMM. These models require segmentation when do recognition, which may incur and propagate the inaccuracy. Our method obtains 0.141, 0.123 and 0.111 higher accuracy, respectively, showing the advantage of circumventing the temporal segmentation.

As presented at the bottom of Tab.~\ref{tab:accu}, we test the schemes that makes up the missing of alignment during recognition, as mention in previous subsection \textit{Learning and Recognition of LS-HAN Model}. We see that LS-HAN with scheme (c) achieves the highest accuracy. Although this alignment is not the true alignment result, we still suppose this scheme is feasible for our model considering the significant results.

\begin{figure}[t]
 \centering
 \includegraphics[width=0.9\linewidth]{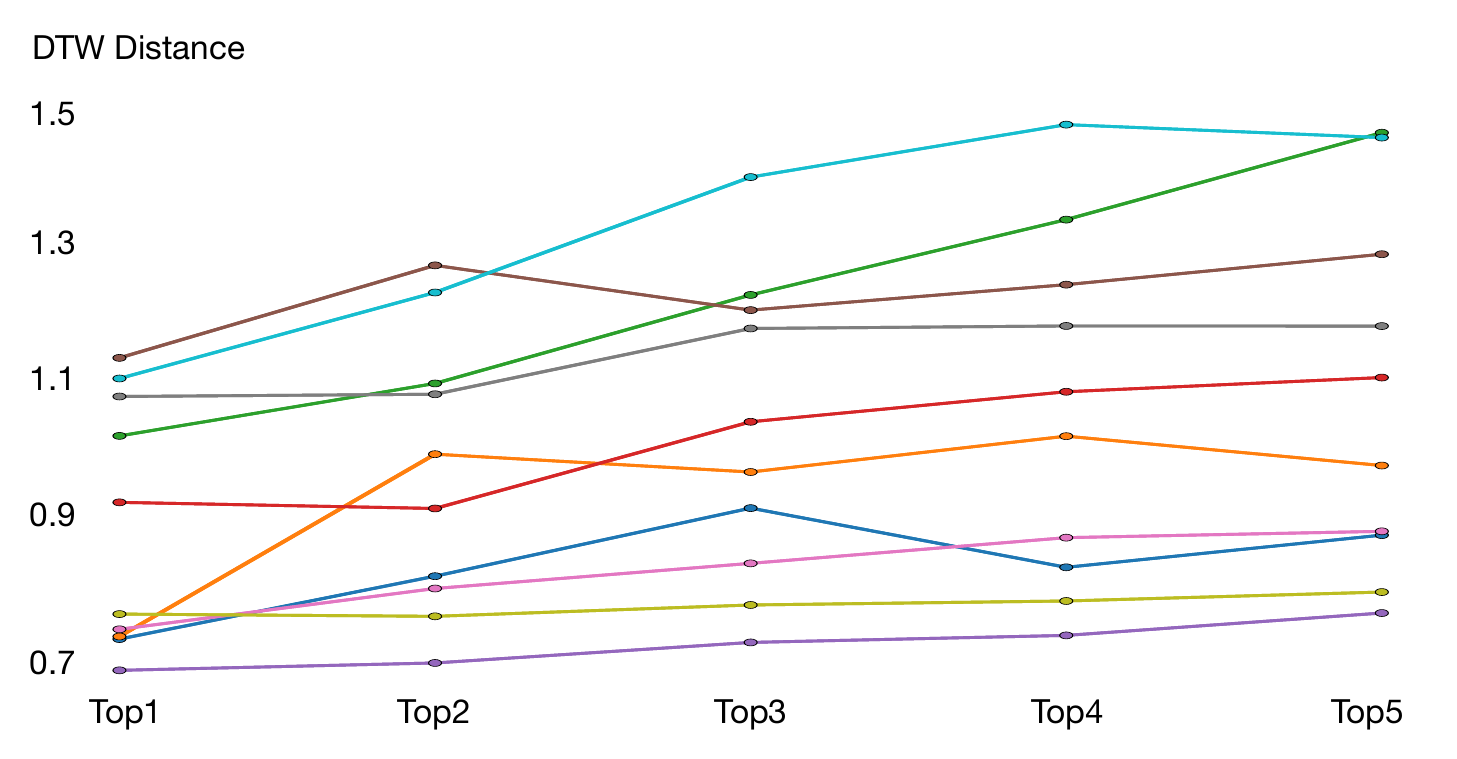}
 \caption{LSTM-based and Latent Space similarity measure comparison.}
 \label{fig:lstm-dtw}
\end{figure}

%
%

Tab.~\ref{tab:accu2} shows a comparison of our result with the recently published results. For a fair comparison, we only utilize hands sequences, as with all competing methods. Both Deep Hand and Recurrent CNN are extensions to CNN. The former combines CNN with an EM algorithm, while the latter proposes the RNN-CNN, which takes advantage of both the feature learning capability of CNN and the temporal sequence modeling capability of the iterative EM and RNN. Our approach shares a similar idea but goes a step further, which bridges semantic gap with a latent space, then applies HAN to hypothesize semantic sentence. The presented LS-HAN outperforms both Deep Hand and Recurrent CNN. 
\subsection{Relationship Between HAN and LS}
In the proposed LS-HAN,  video-sentence correlations are indicated by both the most likely sentence outputs from HAN and the distance metrics in the Latent Space. Theoretically, these two measures should be identical, therefore, an empirical test is carried out and the results are presented in Fig.~\ref{fig:lstm-dtw}. $10$ video sequences are randomly selected and fed into the HAN, with each video generating $5$ most likely sentences. In Fig.~\ref{fig:lstm-dtw}, these $5$ mostly likely sentences are denoted as vertices on a polyline and sorted with descending probabilities along the X-axis. In addition, the DTW distances in the Latent Space between the video (indicated by the polyline) and each sentences (denoted by the vertices) are visualized along the Y-axis. Theoretically, if these two similarity measures are identical, all polylines in Fig.~\ref{fig:lstm-dtw} should be monotonically increasing. Practically there are some noises but the increasing trends are within our expectations.

\begin{table}
	\centering
	\small
	\caption{Continuous SLR on RWTH-PHOENIX-Weather.} \label{tab:accu2}
	\setlength{\tabcolsep}{10pt} %
	\begin{tabular}{l|c} \hline
		Methods &  Accuracy\\ \hline
		\cite{koller2015continuous} &  0.444 \\ 
		Deep Hand \cite{koller2016deep}          &  0.549 \\ 
		Recurrent CNN \cite{cui2017recurrent}        &  0.613 \\
		\textbf{LS-HAN (only hand sequence)}                  &  0.617\\
		\hline
	\end{tabular}
\end{table}

\subsection{Sensitivity Analysis on Parameter Selections}
\begin{figure}[t]
\centering
		\includegraphics[width=0.95\linewidth]{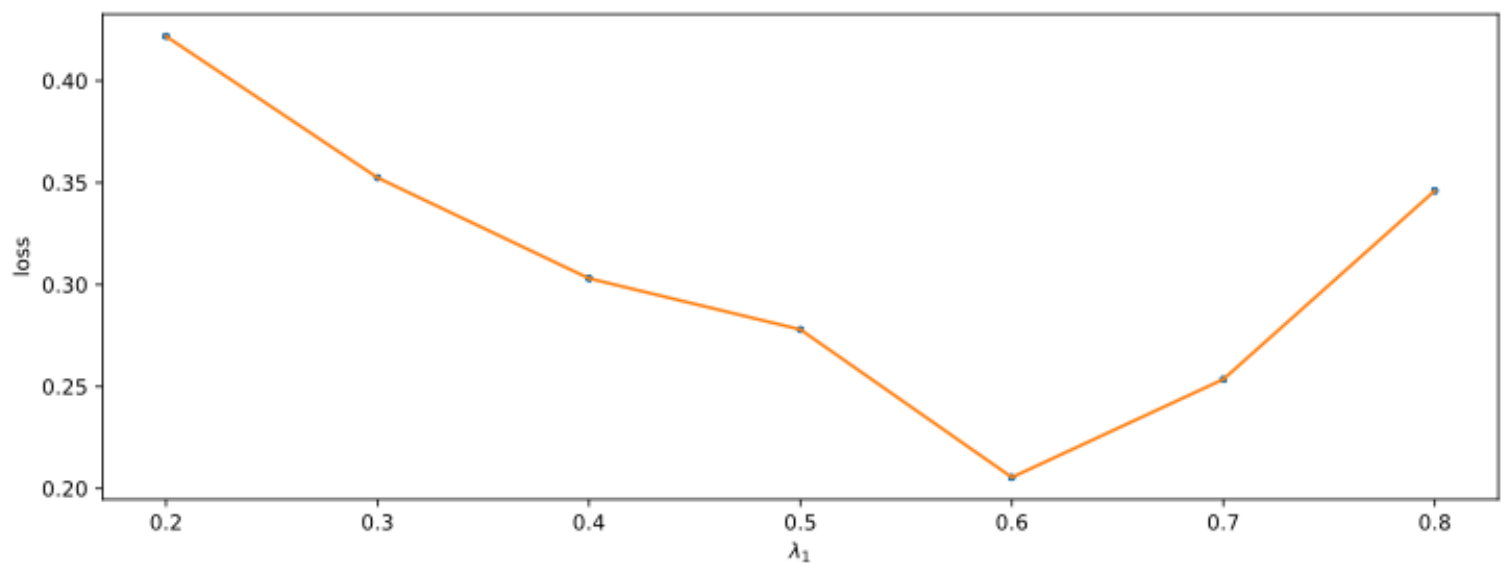}
	\caption{Validation error rate with respect to varying trade-off parameter $\lambda_1$ in Eq.~\eqref{eq:loss}.}
	\label{fig:lambda}
\end{figure}

The robustness of the proposed LS-HAN with respect to different selections of parameters are summarized in this section. A sample sensitivity analysis on the trade parameter $\lambda_1$ in Eq.~\eqref{eq:loss} is presented here and illustrated in Fig.~\ref{fig:lambda}. LS-HAN recognition accuracy is evaluated with 2.5k instances in the validation set with varying $\lambda_1$ values. Expectedly, extreme $\lambda_1$ values lead to high error rate; while the optimal choice is approximately $0.6$. Such reasonable sensitivity behaviors also verify the validity of establishing an HAN model in the video-sentence latent space. 

\section{Conclusion}
\label{sec:con}
In this paper, the LS-HAN framework is proposed for continuous SLR is proposed, which eliminated both the error-prone temporal segmentation and the sentences synthesis in post-processing steps. For video representation, a two-stream 3D CNN generates highly informative global-local features, with one stream focused on global motion information and the other on local gesture representations. A Latent Space is subsequently introduced via an optimization of labeled video-sentence distance metrics. This latent space captures the temporal structures between signing videos and annotated sentences by aligning frames to words.  Our future work could involve the extension of the LS-HAN to longer compound sentences and real-time translation tasks. 

\section{Acknowledgments}
The work of H. Li was supported in part by the 973 Program under Contract 2015CB351803 and in part by NSFC under Contract 61325009 and Contract 61390514. The work of W. Zhou was supported in part by NSFC under Contract 61472378 and Contract 61632019, in part by the Young Elite Scientists Sponsorship Program by CAST under Grant 2016QNRC001, and in part by the Fundamental Research Funds for the Central Universities. This work is partially supported by Intel Collaborative Research Institute on Mobile Networking and Computing (ICRI-MNC). 

\end{document}